\documentclass[10pt, conference]{IEEEtran}
\IEEEoverridecommandlockouts
\usepackage{cite}
\usepackage{amsmath,amssymb,amsfonts}
\usepackage{algorithmic}
\usepackage{graphicx}
\usepackage{textcomp}
\usepackage{xcolor}
\def\BibTeX{{\rm B\kern-.05em{\sc i\kern-.025em b}\kern-.08em
    T\kern-.1667em\lower.7ex\hbox{E}\kern-.125emX}}
\begin{document}

\title{Experience Constrained Hierarchical Federated Reinforcement Learning for Large-scale UAV Teams in Hazardous Environments}

\author{\IEEEauthorblockN{1\textsuperscript{st} Qinwei Huang}
\IEEEauthorblockA{\textit{EECS} \\
\textit{Syracuse University}\\
Syracuse, USA \\
qhuang18@syr.edu}
\and
\IEEEauthorblockN{2\textsuperscript{nd} Rui Zuo}
\IEEEauthorblockA{\textit{EECS} \\
\textit{Syracuse University}\\
Syracuse, USA \\
rzuo02@syr.edu}
\and
\IEEEauthorblockN{3\textsuperscript{rd} Simon Khan}
\IEEEauthorblockA{\textit{AFRL} \\
\textit{Air Force Research Laboratory}\\
Rome, USA \\
simon.khan@us.af.mil}
\and
\IEEEauthorblockN{4\textsuperscript{th} Qinru Qiu}
\IEEEauthorblockA{\textit{EECS} \\
\textit{Syracuse University}\\
Syracuse, USA \\
qiqiu@syr.edu}
}

\maketitle

\begin{abstract}
Conventional federated learning assumes that greater learner participation improves training performance, by leveraging abundant, independently generated local data. However, in federated reinforcement learning (FRL) for unmanned aerial vehicle (UAV) teams in hazardous environments where experience generation is severely constrained by safety considerations, energy limitations, and mission duration, this assumption may break. This work introduces Experience-Constrained Hierarchical Federated Reinforcement Learning (EC-HFRL), a framework in which clusters act as federated learning agents, while multiple intra-cluster learners represent parallel learning resources that reuse a shared experience pool. We show that increasing participation does not necessarily improve learning performance. Instead, learning performance is strongly associated with experience reuse strategy and the dominance of key analytically identified gradient transition experiences within a cluster. In particular, minibatch size primarily determines effective replay exposure, while higher intra-cluster participation increases reuse level. Empirical results demonstrate that the performance regimes are strongly associated with the structure of the learning signal, rather than federated aggregation effects, clarifying the limited and secondary role of learner participation in experience-constrained FRL.
\end{abstract}

\begin{IEEEkeywords}
federated reinforcement learning, experience-constrained learning, prioritized experience replay, multi-UAV systems
\end{IEEEkeywords}

\section{Introduction}
Unmanned aerial vehicles (UAVs) are increasingly deployed for missions in hazardous environments, including radiation inspection in contaminated areas, wildfire response, and post-disaster structural assessment \cite{Chung2018SwarmSurvey}. In such settings, agents must operate under severe uncertainty in environmental dynamics and hazard distributions, such as toxic gas dispersion and radiation exposure \cite{hutchinson2019source}. Although deep reinforcement learning provides a principled framework for adaptive control, real-world deployment in hazardous environments introduces a fundamental and often overlooked constraint: safe and informative experience can only be generated at a limited rate.

Unlike conventional multi-agent learning settings where large-scale data collection is feasible, experience generation in hazardous environments is inherently slow, costly, and risky. Each task execution carries a non-negligible probability of irreversible UAV failure due to environmental exposure, collision, or communication loss. Moreover, hazards are typically unknown and only partially observable: UAVs have no prior knowledge of hazard sources, spatial extent, or failure mechanisms, and cannot directly observe when or how exposure leads to failure. As a result, agents must infer hazardous regions indirectly from onboard sensing, sparse rewards or penalties, and observed failures while learning to avoid destructive conditions under severe uncertainty.

These constraints substantially alter the learning regime in experience-constrained FRL settings. UAVs are not freely reusable training entities, and each deployment consumes a limited physical resource. Consequently, the total amount of experience that can be safely generated is bounded by operational constraints rather than computational capacity. Even with large fleets that have abundant parallel computation resources, learning is still constrained by a low experience generation rate. At the same time, 
participation in learning and communication incurs non-negligible energy costs for each agent.
In this regime, increasing the number of parallel learners does not expand the available data pool; instead, it reshapes how a limited experience pool is repeatedly used during optimization. As a result, architectural choices that increase parallel learning participation may significantly increase system-level energy consumption, with marginal gains in learning performance.

Federated reinforcement learning (FRL) provides a natural paradigm for collaborative learning among edge devices by enabling decentralized policy optimization with limited data sharing \cite{qi2021federated,wang2023federated}. Most existing FRL methods, however, implicitly assume abundant training data and emphasize scalability through increased participation or efficient aggregation\cite{nishio2019client,zhang2024device}. Under fixed experience budgets, these assumptions no longer hold: learning dynamics are shaped by finite-sample effects, experience reuse, and the structure of stochastic updates, altering the role of participation during training.

In this work, we study learning dynamics in FRL under constrained experience generation using a cluster-based hierarchical framework \cite{wang2022accelerating}. We focus on \emph{experience-constrained federated reinforcement learning}, where the rate of safe experience generation is fixed by environmental, safety, or mission constraints. In this regime, increasing learner participation changes how a fixed experience pool is reused rather than introducing additional data. Our analysis is restricted to this fixed-budget setting rather than data-abundant or throughput-oriented reinforcement learning scenarios.


We propose a hierarchical architecture for Experience-Constrained Hierarchical Federated Reinforcement Learning (EC-HFRL) where UAVs are organized into clusters with no inter-cluster experience sharing. In each training round, only a small subset of UAVs within each cluster interacts with the hazardous environment to generate new experience, while the remaining UAVs learn from these cluster-specific experiences through local updates. The local models are first aggregated at the cluster level, and then aggregated at the system level. Our goal is to manage these learning resources to maximize learning performance under fixed experience budgets, while analyzing the resulting energy implications of different participation configurations.

From a learning-signal perspective, we analyze how the parameters of this architecture shape optimization. Increasing the number of intra-cluster learners introduces more concurrent updates from the same replay pool but does not increase experience availability. Their participation therefore primarily affects the statistical stability of replay sampling. On the other hand, minibatch structure, particularly batch size, determines which transitions dominate the loss and shape the policy updates. Under constrained experience budgets, learning progress depends strongly on whether semantically critical transitions, referred to as \emph{key experiences}, enter the loss-dominant subset of prioritized replay. 
When such transitions are absent, increased reuse and aggregation amplify uninformative gradients, leading to diminished returns despite higher cost.

We validate these findings on two large-scale UAV team tasks modeling chemical leakage detection and wildfire hazard assessment, where agents learn under unknown and potentially destructive environmental conditions. Across both domains, we observe consistent performance regimes driven by learning-signal structure under fixed experience budgets. These results demonstrate that performance variation in data-scarce FRL is strongly associated with learning signal composition rather than participation intensity, while participation density primarily governs system-level energy cost. 

The main contributions of this paper are summarized as follows:
\begin{itemize}
    \item We propose EC-HFRL, a hierarchical FRL under \emph{fixed per-round experience budgets}, 
    designed for multi-UAV deployments where hazardous conditions fundamentally limit safe experience generation.

    \item We analyze how intra-cluster learning resource allocation and minibatch configuration shape learning dynamics under this regime, showing that increased participation primarily stabilizes replay sampling, while minibatch structure determines which transitions dominate optimization. 

    \item We identify two system metrics, Key Enrichment Ratio (KER) and Key TD Contribution (KTC), as the indicators of the contribution of \emph{key experiences} in the learning process, and demonstrated that they are strongly associated with effective learning. 

    \item Through empirical studies of EC-HFRL on UAV team tasks in two hazardous-environment tasks, we analyze learning-signal structure and reveal how $(K,b)$ shapes  learning performance and energy cost under fixed experience budgets.
\end{itemize}

\section{Related Work}

\subsection{Experience-Constrained Reinforcement Learning and Replay}

Many real-world reinforcement learning problems operate under severe data constraints, where experience generation is costly, risky, or rate-limited. Experience replay is widely adopted to improve sample efficiency by reusing previously collected transitions, and prioritized experience replay (PER) further biases sampling toward transitions with large TD errors to accelerate learning under limited data \cite{schaul2015prioritized}.

Most replay-based methods focus on designing sampling priorities for individual transitions and analyzing their effects on convergence or stability. However, these approaches typically place less emphasis on how learning signals are aggregated within minibatches, especially under repeated replay reuse. When minibatches are large, task-critical transitions can be diluted by gradient averaging, weakening their influence on updates.

This limitation becomes more pronounced when replay is combined with parallel learning from a shared buffer with overlapping samples. While parallelism is often assumed to improve learning efficiency, existing work rarely examines its interaction with replay under fixed experience budgets, where parallelism primarily intensifies reuse.

Our work addresses this gap in experience-constrained federated reinforcement learning by analyzing replay exposure and minibatch structure under constrained experience generation, and by characterizing how these factors jointly determine which transitions dominate learning signals during optimization.



\subsection{Federated and Parallel Reinforcement Learning under Limited Data}

Federated learning (FL) enables distributed optimization through parallel local updates and periodic parameter aggregation \cite{mcmahan2017communication,lim2020federated}. In practice, FL systems often operate under partial participation due to communication, computation, or availability constraints \cite{nishio2019client}. In supervised learning, increased participation is commonly associated with improved convergence because additional clients typically contribute independent data.

In hierarchical FRL deployments for physical multi-agent systems, experience generation is often limited by safety and mission constraints. In these regimes, increasing learner participation mainly changes how a fixed experience budget is reused across updates. Despite this distinction, much of the existing FRL literature still adopts participation intuitions inherited from data-abundant FL settings.

Parallel reinforcement learning architectures such as A3C \cite{mnih2016asynchronous} and IMPALA \cite{espeholt2018impala} leverage parallelism to improve throughput in experience collection and policy optimization. Prior work has shown that increased parallelism can introduce policy lag, correlated updates, and finite-sample effects, leading to diminishing returns unless explicitly addressed. These issues are exacerbated when parallel learners repeatedly sample from a shared replay buffer, yet their interaction with participation density remains underexplored. Unlike throughput-oriented parallel RL settings, our focus is on regimes where parallelism only redistributes reuse of a fixed experience budget.

Recent studies in federated optimization further indicate that participation patterns influence convergence behavior beyond communication efficiency \cite{wang2022unified}. In contrast to system-oriented analyses emphasizing scheduling or aggregation efficiency, our work focuses on federated reinforcement learning under fixed per-round experience budgets and analyzes how learner participation and minibatch configuration shape learning behavior through experience reuse and learning signal composition.

\section{Problem Formulation}

We consider federated reinforcement learning for large-scale multi-UAV systems operating under strict energy constraints and limited experience generation. The learning problem is formulated as a decentralized partially observable Markov decision process (Dec-POMDP) \cite{oliehoek2016concise}, augmented with explicit constraints on energy consumption and experience availability. Our objective is to characterize how learning dynamics depend on experience aggregation and learner resource allocation when data generation is inherently bounded by hazardous operating conditions. Here, a cluster $c$ is a federated learning entity. We use $\mathbb{C}_c^t$ and $\mathbb{K}_c^t$ to denote the set of UAVs in cluster $c$ and the set of UAVs participating learning in $c$ at time $t$, $\mathbb{K}_c^t\subseteq{\mathbb{C}_c^t}$. $\mathbb{K}_c^t$ is randomly selected over time to ensure an even distribution of the workload and UAVs in $\mathbb{K}_c^t$ are parallel learning resources operating on a shared experience pool.

\subsection{Resource-Constrained Multi-UAV Model}

We consider a group of $N$ UAVs indexed by $i \in \{1,\dots,N\}$ operating in a bounded three-dimensional workspace $\mathcal{X} \subseteq \mathbb{R}^3$. Each UAV is subject to strict SWaP (Size, Weight, and Power) constraints, such that both mission execution and learning-related activities incur non-negligible energy costs.

The global system state at time $t$ is defined as
\[
s^t = [\mathbf{P}^t, \mathbf{V}^t, \mathbf{E}^t, \mathcal{H}^t],
\]
where $\mathbf{P}^t$ and $\mathbf{V}^t$ denote the joint positions and velocities of all agents, $\mathbf{E}^t = \{e_1^t,\dots,e_N^t\}$ represents residual onboard energy levels, and $\mathcal{H}^t$ denotes latent environmental hazard parameters that influence transition dynamics and failure risk.

Energy consumption arises from both physical operation and learning participation. Physical execution consumes energy through propulsion and control, while learning incurs additional energy cost due to local policy optimization and model communication. The energy dynamics of agent $i$ are given by
\begin{equation}
e_i^{t+1} = e_i^t
- \mathbb{I}(\text{mode}_i=\text{Fly}) \cdot E_{\mathrm{fly}}(a_i^t)
- \mathbb{I}(i \in \mathbb{K}^t) \cdot E_{\mathrm{learn}},
\end{equation}
where $\mathbb{K}^t$ denotes the set of agents participating in learning during communication round $t$. 

The learning-related energy cost is decomposed as
\[
E_{\mathrm{learn}} = E_{\mathrm{train}} + E_{\mathrm{comm}} + E_{\mathrm{agg}},
\]
where $E_{\mathrm{train}}$ accounts for local optimization computation, $E_{\mathrm{comm}}$ captures wireless model transmission and reception, and $E_{\mathrm{agg}}$ represents aggregation and broadcast overhead at the cluster head. All energy terms are computed using a physics-grounded power model by integrating device-specific power consumption over execution time, with parameters derived from representative embedded compute and radio modules used in UAV platforms.

This formulation captures the coupling between mission execution, learning participation, and finite energy budgets, enabling quantitative analysis of the energy implications of different participation and minibatch configurations. 
Energy modeling is used exclusively for post-hoc evaluation of deployment-level trade-offs and does not alter the learning objective, reward function, or policy optimization process.

\subsection{Hazard-Aware Environment Dynamics}

Environmental risk is modeled by a continuous hazard field $\mathcal{Z}: \mathcal{S} \rightarrow [0,1]$ defined over the environment state space $\mathcal{S}$. For an agent at state $s_t \in \mathcal{S}$, the hazard field specifies local risk intensity, which is mapped to the probability of catastrophic failure via a monotonic function $\mathcal{F}(\cdot)$. Specifically, failure at time $t$ is sampled as
\begin{equation}
P_{\mathrm{fail}}(s_t) \sim \mathrm{Bernoulli}(\mathcal{F}(\mathcal{Z}(s_t))).
\end{equation}

This failure mechanism induces irreversible termination and directly constrains experience generation by limiting the duration and frequency of safe interactions. The hazard field may depend on spatial location or jointly on agent state variables, leading to heterogeneous transition dynamics across tasks. These differences give rise to distinct learning regimes, which are examined empirically in Sec.\ref{sec:exp_env}. The hazard model affects transition dynamics and data availability but does not introduce explicit safety constraints or risk-sensitive optimization objectives.

\subsection{Learning Participation under Fixed Experience Budgets}

We consider a cluster-based federated reinforcement learning architecture in which UAVs are partitioned into clusters for hierarchical optimization.
Each cluster is treated as a federated learning entity and maintains its own experience pool, while clusters do not share experience with one another.

Within each cluster $c$, a fixed-size subset of UAVs is designated as active learners, denoted by $\mathbb{K}_c \subset \mathcal{C}_c$ with $|\mathbb{K}_c^t| = K^t$.
These active learners act as parallel learning resources and perform local policy updates by reusing cluster-specific experience. The aggregated update at communication round $t$ is given by
\begin{equation}
\mathbf{g}_t = \sum_{i \in \mathbb{K}_c^t} \nabla \mathcal{L}_i(\theta).
\end{equation}
This algorithmic formula illustrates aggregated local updates and does not assume a specific optimizer. Learning-rate scaling and normalization are omitted as our analysis focuses on relative learning-signal composition.

A defining characteristic of physical multi-agent systems is that the amount of newly generated experience per communication round is constrained by mission duration, safety considerations, and energy availability. Accordingly, we assume that each cluster contributes a fixed amount of experience per round, independent of the number of participating learners. Active learners sample minibatches from a shared cluster-level experience pool to perform local updates and the sampling is performed independently to save communication.


This formulation establishes a controlled setting for analyzing learning dynamics under experience scarcity. Participation size $K$ and minibatch size $b$ act as external configuration parameters that shape experience reuse and gradient aggregation, rather than decision variables optimized by the learning algorithm, and are fixed throughout each experimental run.

\begin{figure}[t]
  \centering
  \includegraphics[width=\columnwidth]{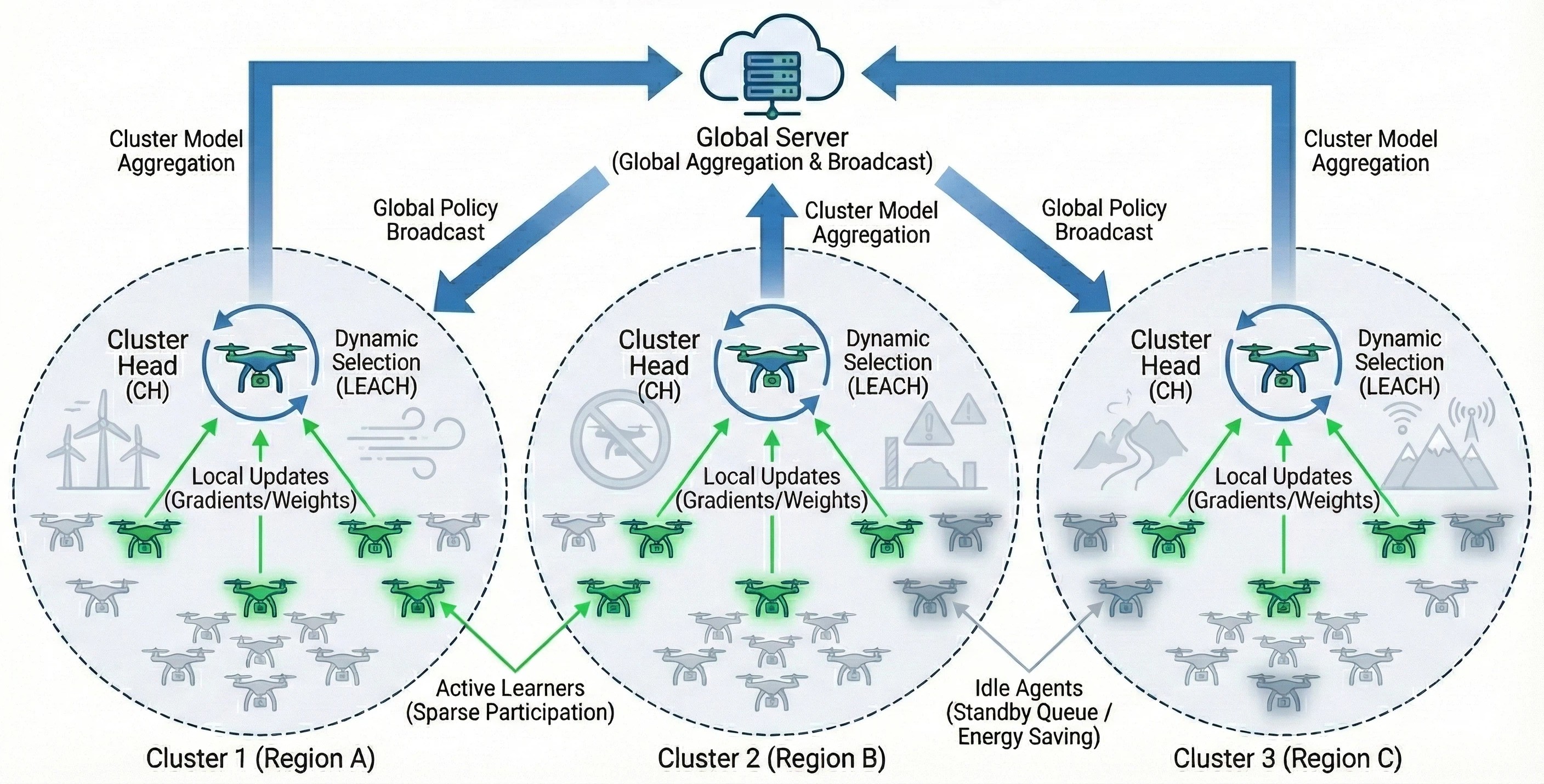}
  \caption{Overview of the EC-HFRL framework.}
  \label{fig:framework}
\end{figure}

\section{Method}
\label{sec:method}

This section presents the EC-HFRL training protocol. 
We characterize prioritized replay under multi-learner sampling and analyze how learner participation and minibatch configuration shape replay exposure during optimization. These relationships provide the structural foundation for our subsequent empirical analysis of which exposed transitions are associated with downstream performance.

\subsection{Training Protocol under Fixed Experience Budgets}
\label{sec:protocol_fixed_budget}

EC-HFRL adopts a cluster-based hierarchical federated reinforcement learning architecture, illustrated in Fig.\ref{fig:framework}. UAV agents are partitioned into geographically or operationally defined clusters. Within each cluster, a designated cluster head (CH) \cite{1045297} aggregates model updates from participating agents, while a global server periodically aggregates cluster-level models and broadcasts the updated policy to all clusters. CH rotation follows the LEACH protocol \cite{1045297}, which probabilistically rotates the CH role to balance energy consumption across UAVs.
This hierarchy enables scalable coordination across large UAV teams and localizes experience storage and replay at the cluster level, serving as a system-level organization for distributed learning without modifying the underlying reinforcement learning algorithm.

Training proceeds in synchronized communication rounds. In each round and for each cluster $c$, only a small subset of UAVs interacts with the environment to collect new trajectories, which are stored in a shared cluster-level replay buffer $\mathcal{D}_c^t$. The amount of newly collected experience per round is rate-limited by hazard-driven operational constraints, such as mission duration and recovery cycles, and is therefore treated as fixed. 

Within each cluster, a subset of agents is selected as \emph{active learners} $\mathbb{K}_c^t$, with $|\mathbb{K}_c^t| = K$. Here, each cluster is treated as a federated learning agent, while the active learners $\mathbb{K}_c^t$ represent parallel learning resources allocated within the cluster that reuse a shared experience pool. Each active learner independently samples minibatches of size $b$ from the logically shared replay buffer to perform local policy updates. Local updates are aggregated at the cluster head and subsequently synchronized at the global server using standard average aggregation. Aggregation follows standard federated averaging and is not modified or adapted based on replay statistics or learning signal properties.

Under this protocol, learner participation $K$ does not affect the rate of experience generation. Instead, the configuration $(K, b)$ shapes how a fixed experience pool is sampled and reused across parallel stochastic updates. Throughout this work, $K$ and $b$ are treated as externally configured parameters rather than algorithmic decision variables, forming the basis for the replay exposure analysis in the next subsection.

\subsection{Prioritized Replay under Multi-Learner Sampling}
\label{sec:method_replay_exposure}

Under the protocol described above, we analyze prioritized replay when multiple learners concurrently sample from the same local replay buffer. In this setting, prioritized replay not only determines how frequently individual transitions are reused, but also how much distinct experience is effectively exposed to learning updates within a communication round.

To characterize this effect, we introduce a replay exposure measure that quantifies the distinct transition support covered by the union of sampled minibatches. Formally, let $\mathcal{D}$ denote the replay buffer in a given round with $|\mathcal{D}| = N$ transitions. Each learner $k \in \{1,\dots,K\}$ samples a minibatch $\mathcal{M}_k$ of size $b$ according to its local prioritized sampling distribution. Let $\left|\bigcup_{k=1}^{K}\mathcal{M}_k\right|$ denote the number of \emph{distinct} transitions, identified by buffer indices, that appear at least once across all learners. We define the \emph{union coverage ratio} (UCR) as
\begin{equation}
\label{eq:ucr}
\mathrm{UCR} \;\triangleq\; \frac{\left|\bigcup_{k=1}^{K} \mathcal{M}_k \right|}{K b}.
\end{equation}
Repeated samples of the same transition across learners do not increase the numerator, so a smaller UCR indicates stronger concentration and greater overlap under a fixed sampling budget. We emphasize that UCR does not measure replay buffer coverage, but rather the distinctness of replayed transitions per unit sampling budget within a communication round.

This definition makes explicit how replay exposure exhibits a clear association with both parallelism and minibatch structure. Under a highly concentrated priority distribution, many draws fall on a small subset of high-priority transitions. As $K$ increases with $b$ fixed, the total number of draws $K b$ increases, but the number of distinct transitions grows sublinearly due to repeated sampling of the same high-TD replay subset. Consequently, the number of distinct transitions exposed to learning grows much more slowly than the total number of draws, and the UCR decreases. Similarly, larger minibatches amplify concentration effects through gradient averaging, further reducing the effective diversity of samples contributing to updates.

Importantly, this exposure compression arises from the interaction between the concentration of the replay priority distribution and parallel reuse from a logically shared replay distribution, rather than from stochastic variability alone. While $b$ controls per-learner sampling granularity, $K$ shapes the extent of within-round reuse over the same high-TD replay subset. Replay exposure under multi-learner prioritized sampling is therefore jointly shaped by $(K,b)$ and the distribution of replay priorities.

\subsection{Key Experience Replay under Constrained Exposure}
\label{sec:method_key_experience}
Repeated sampling of high-priority transitions alone does not guarantee effective learning under constrained replay exposure.   Transitions that steer policy updates toward task success may constitute only a small fraction of the replay buffer. We refer to such transitions as \emph{key experiences}. Here, key experiences are introduced as an analytical notion to reason about learning relevance under constrained exposure, rather than as algorithmic entities available during training. In particular, under fixed buffer capacity and compressed exposure, such key experiences that contribute meaningfully to learning progress are even more limited. 

In hazardous environments with asymmetric reward structures, many transitions provide limited or redundant gradient information, while a small subset consistently shapes the direction of policy updates. Which transitions are informative is closely linked to both the task and the current policy state.

Importantly, key experiences are identified based on learning relevance as assessed post hoc,  rather than outcome of the rewards. Depending on the task and training stage, they may correspond to successful behaviors, near-boundary avoidance events, or failure cases that reveal hazardous conditions. As training progresses, the semantic role of key experiences shifts with changes in exploration patterns, error modes, and remaining uncertainty. Consequently, key experiences are neither static nor uniformly distributed across the replay buffer. Conceptually, we distinguish two broad learning stages when interpreting key experiences. In the early stage, the dominant bottleneck is to discover a policy that can reliably reach the destination region or complete the task. In the later stage, once such behavior becomes stable, the bottleneck shifts to refinement, such as avoiding hazard, reducing unnecessary failure, or improving safety and efficiency. Accordingly, the criteria used to identify key experiences are stage-dependent and may change as the dominant learning bottleneck shifts over training.

Prioritized replay increases the sampling frequency of transitions with large temporal-difference (TD) errors. 
However, TD magnitude alone does not determine learning usefulness. Large TD errors may also arise from instability, stochasticity, or failure modes with limited long-term value, allowing uninformative transitions to dominate replay usage without contributing to task-relevant learning.

Under the multi-learner sampling protocol described in Sec.\ref{sec:method_replay_exposure}, independent sampling from learners further concentrates replay exposure on the portion of transitions with high TD error. 
Only transitions that appear in the union of sampled minibatches across learners are eligible for subsequent amplification, effectively defining an admissible replay support. Replay exposure refers to the set of unique transitions that are sampled into the minibatch. It therefore acts as an effective filter: it influences which transitions are repeatedly reused and which remain unseen, regardless of their potential learning relevance.

These observations motivate a shift from analyzing replay intensity to analyzing which transitions are actually exposed and reused during training.
Rather than focusing solely on aggregated TD magnitude, we emphasize the presence of key experiences within the exposed replay support. In the following sections, we empirically examine how exposure compression influences the admission of key experiences over training and how their presence is empirically associated with downstream performance.

\section{Experiments}

We design experiments to examine how learner participation and minibatch configuration shape learning dynamics in federated reinforcement learning under fixed per-round experience budgets. Rather than comparing algorithmic variants, our experiments isolate how experience reuse, update reliability, and learning signal structure interact across hazardous environments with heterogeneous risk characteristics.

\subsection{Experimental Setup}
\label{sec:exp_env}

To examine how learner participation and minibatch structure interact with experience characteristics under fixed per-round budgets, we evaluate EC-HFRL on two high-fidelity 3D simulation benchmarks with distinct hazard structures. These environments induce qualitatively different distributions of task-critical transitions, enabling a controlled study of how replay exposure and learning-signal composition shape performance across contrasting learning regimes.

\textbf{Position-Dependent Chemical Leak (PD-Chem).}
PD-Chem models a hazardous gas leakage scenario in which environmental risk is determined primarily by agent position relative to the leak source. Early in training, many approach trajectories yield similar reward and exposure patterns, resulting in high semantic redundancy. As destination-reaching behavior becomes reliable, the main bottleneck shifts to refining trajectories by reducing unnecessary exposure and other suboptimal approach behaviors. Accordingly, key experiences in PD-Chem are mainly corrective transitions that distinguish safer and more efficient approaches from poor terminal outcomes.

\textbf{State-Dependent Wildfire Interception (SD-Fire).}
SD-Fire simulates a high-speed wildfire interception task with second-order dynamics, where agents control acceleration under strong inertial effects. Hazard depends jointly on position and velocity, so unsafe outcomes are closely tied to motion-dependent behaviors such as hovering, prolonged exposure, and delayed maneuvering. Because successful interception remains difficult through most of training, the main bottleneck is discovering and stabilizing successful or near-successful interception behavior. Accordingly, key experiences in SD-Fire are mainly success- and near-success-related transitions that provide informative gradients for interception and safe maneuvering.

While the basic concept of key experiences remains unchanged across environments, their semantic composition differs: in SD-Fire, key transitions are predominantly success- or near-success-related experiences, whereas in PD-Chem they more often correspond to corrective or avoidance-related events.

Together, these benchmarks illustrate different dominant optimization bottlenecks under fixed per-round experience budgets.
As a result, learning dynamics in SD-Fire are more sensitive to replay exposure and sample composition.

\begin{figure}[t]
  \centering
  \includegraphics[width=\columnwidth]{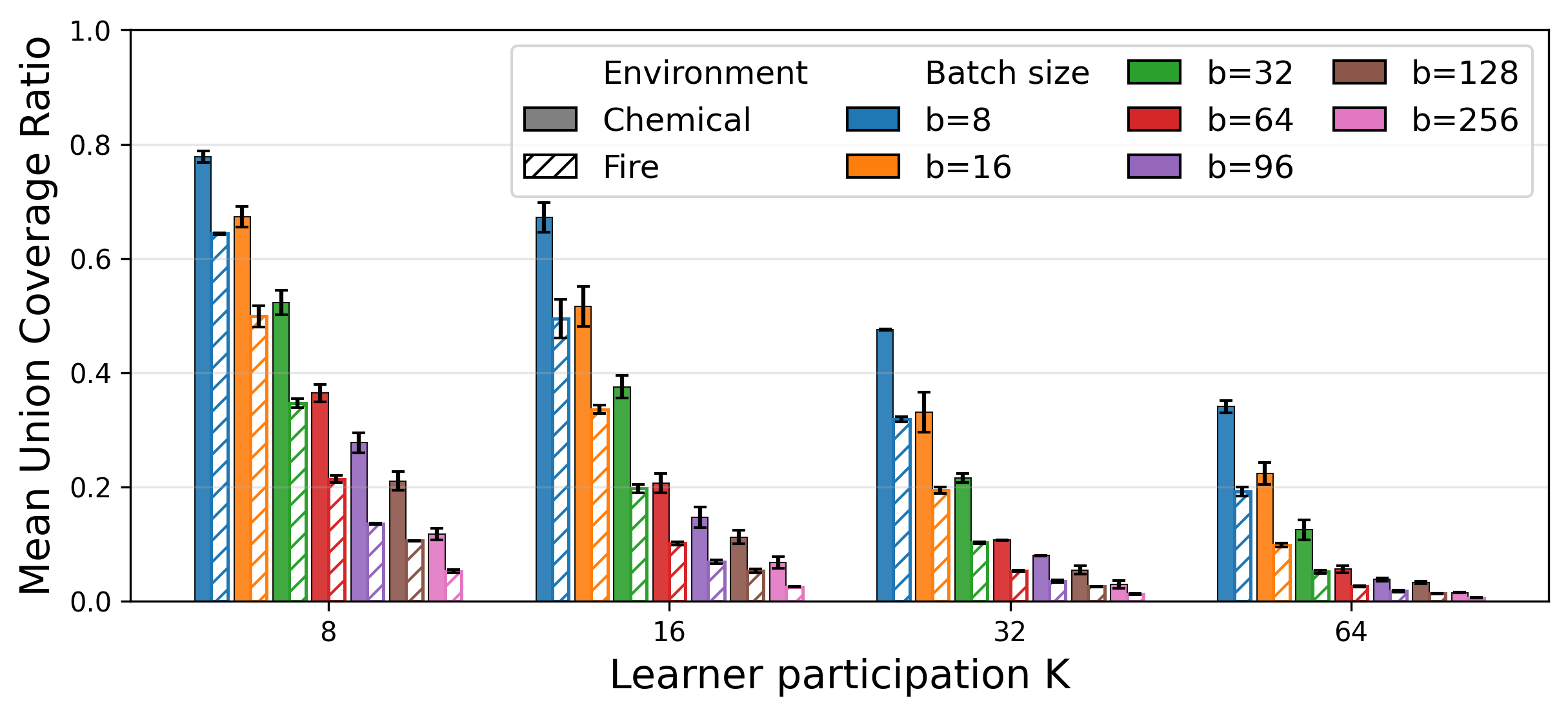}
  \caption{
UCR under different learner participation levels $K$ and minibatch sizes $b$ in the Chemical and Fire environments.
}

  \label{fig:ucrkb}
\end{figure}

\subsection{Replay Exposure under Fixed Experience Budgets}
\label{sec:ucrkb}

We first analyze how learner participation and minibatch structure shape replay exposure under fixed per-round experience budgets. Replay exposure is quantified using the UCR, defined in Eq.\ref{eq:ucr}. Figure\ref{fig:ucrkb} reports the mean UCR under varying learner participation levels $K$ and minibatch sizes $b$ for both environments, revealing consistent trends across tasks.

Under the fixed experience budget setting, increasing learner participation is empirically observed to induce a monotonic decrease in UCR for a fixed minibatch size. As more learners independently sample from a logically shared replay buffer under the same prioritized distribution, sampling overlap increases and replay usage concentrates on a smaller subset of transitions. Minibatch size exhibits a consistently stronger effect. As $b$ increases, UCR decreases sharply across all values of $K$, indicating accelerated concentration of replay usage and a reduced effective support of prioritized replay. 

These results indicate that under fixed experience budgets, replay exposure is shaped in a predictable and monotonic manner by $(K,b)$. Minibatch structure plays a dominant role: quantitatively, increasing minibatch size from $b=8$ to $b=256$ reduces UCR by more than $60\%$ on average across all configurations. In contrast, increasing learner participation from $K=8$ to $K=64$ yields a smaller but consistent reduction in UCR, reflecting increased overlap induced by parallel reuse rather than expanded data coverage. However, reduced replay exposure alone is insufficient to account for learning outcomes. Instead, performance depends critically on the semantic composition of the prioritized samples admitted under this constrained exposure.


\begin{figure}[t]
  \centering
  \includegraphics[width=\columnwidth]{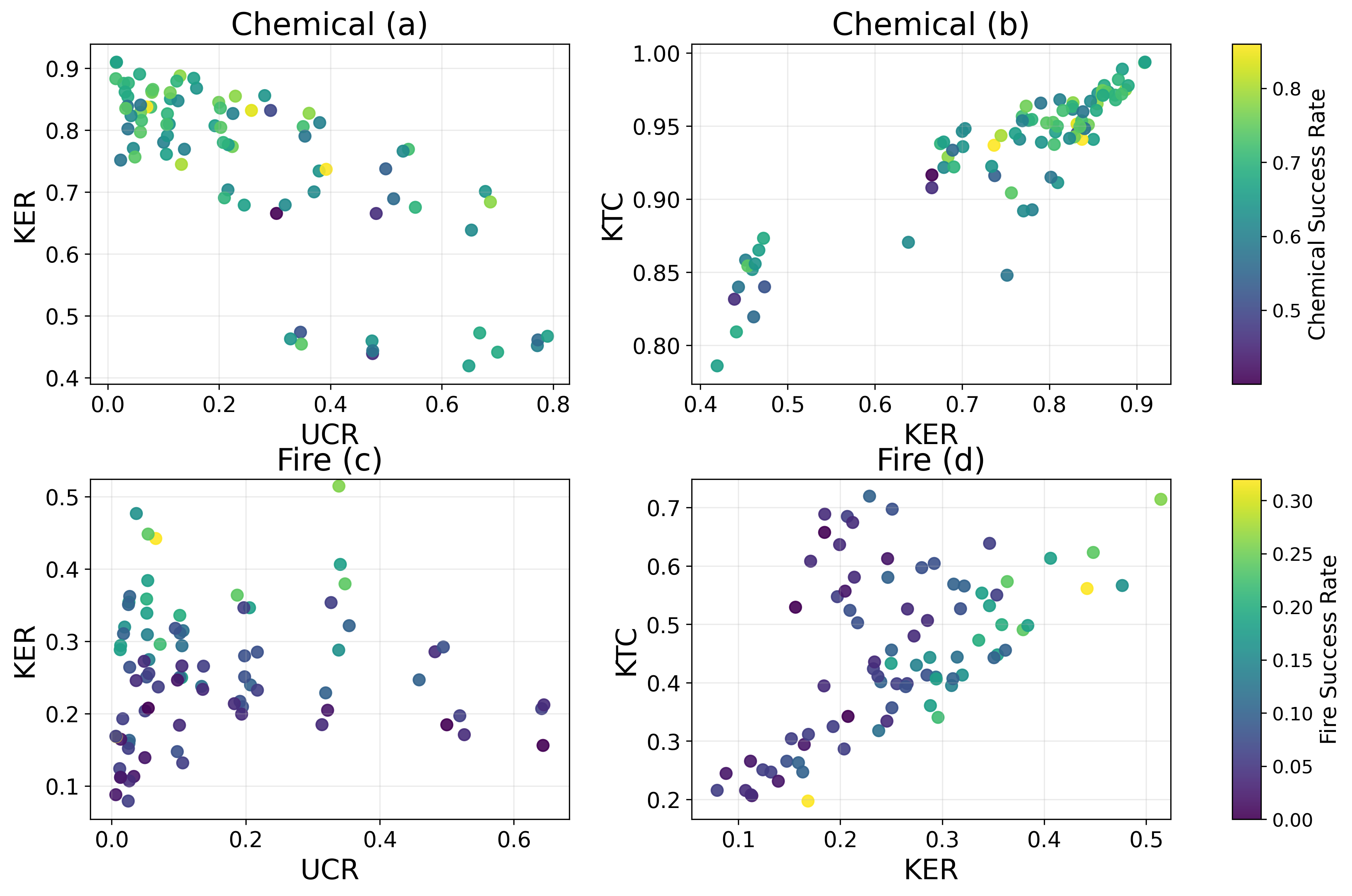}
  \caption{
Replay exposure and key experience diagnostics under fixed experience budgets.
(a)(b) show results for PD-Chem and (c)(d) for SD-Fire.
Left: relationship between replay exposure (UCR) and key experience admission (KER).
Right: relationship between key experience admission (KER) and key TD contribution (KTC).
}

  \label{fig:ucr}
\end{figure}

\subsection{Key Experience Admission and Learning Signal Composition}
\label{sec:sr}

We next examine how replay exposure, quantified by UCR, aligns with changes in the \emph{semantic composition} of prioritized samples and how this composition is reflected in learning signal allocation. Specifically, we analyze whether transitions that define the dominant optimization bottleneck at a given training stage, referred to as \emph{key experiences}, are admitted into the high-TD replay subset and whether they subsequently dominate gradient updates. To this end, we introduce two complementary diagnostics that capture key experience admission at the data level and learning signal allocation at the optimization level.

To assess the role of prioritized replay under fixed experience budgets, we compare PER against uniform replay while holding \(K\) and \(b\) fixed. In PD-Chem, PER improves success rates by 8\%--15\% across minibatch sizes. In SD-Fire, the gap is substantially larger: PER achieves 2--5\(\times\) higher success rates than uniform replay under identical configurations, while uniform sampling often fails to surface success or near-success transitions. These results show that prioritized replay is more effective than uniform sampling at surfacing learning-relevant transitions under fixed experience budgets. We therefore next ask a finer-grained question: among the transitions emphasized by replay, are the task-relevant \emph{key experiences} represented in the high-TD replay subset, and how much TD-error mass do they account for?

\textbf{Key Enrichment Ratio (KER).}
Let \(\mathcal{D}\) denote the replay buffer. For analysis, we define \(\mathcal{T}_p\) as the \emph{high-TD replay subset}, i.e., a compact representation of the average high-TD replay region induced by repeated replay resampling at a fixed checkpoint, with \(p=5\) in our experiments. Given a stage-dependent key indicator \(\kappa(\tau)\in\{0,1\}\), we define

\begin{equation}
\mathrm{KER}
\triangleq
\Pr\!\big(\kappa(\tau)=1 \mid \tau \in \mathcal{T}_p \big)
-
\Pr\!\big(\kappa(\tau)=1 \mid \tau \in \mathcal{D} \big).
\end{equation}
The indicator $\kappa(\tau)$ is constructed post hoc for analysis based on task-specific criteria, and is not available during training. A positive KER indicates that key experiences are overrepresented in the high-TD replay subset relative to the replay buffer as a whole. KER captures enrichment in the high-TD replay subset without weighting transitions by TD magnitude. In implementation, KER is estimated by averaging over replay-resampled minibatches; \(\mathcal{T}_p\) is used as compact notation for the resulting average high-TD replay region at a fixed checkpoint.

Figure\ref{fig:ucr}(a,c) plots KER against UCR for both environments, with each point corresponding to a $(K,b)$ configuration and color indicating test success rate. In PD-Chem, lower UCR configurations tend to exhibit higher KER, suggesting that replay becomes more concentrated on stage-relevant corrective transitions once successful behaviors become common. In SD-Fire, the relationship between UCR and KER is less monotonic: replay exposure alone does not guarantee that key experiences are sufficiently represented in the high-TD replay subset, leading to wide variability in KER at similar UCR levels. 

To assess whether key experiences also account for a large share of TD-error mass within the high-TD replay subset, we further examine their relative TD contribution.
\textbf{Key TD Contribution (KTC).}
We define
\begin{equation}
\mathrm{KTC}
\triangleq
\frac{
\sum_{\tau \in \mathcal{T}_p}
\mathbb{I}[\kappa(\tau)=1]\;|\delta(\tau)|
}{
\sum_{\tau \in \mathcal{T}_p}
|\delta(\tau)|
},
\end{equation}
where $\delta(\tau)$ denotes the TD error of transition $\tau$. KTC measures the fraction of absolute TD error within the high-TD replay subset contributed by key experiences and serves as a proxy for their relative contribution to gradient updates.

Figure\ref{fig:ucr}(b,d) shows a strong positive association between key experience admission and TD-mass contribution. Configurations with higher KER tend to have higher KTC, indicating that key experiences account for a larger fraction of TD-error mass within the high-TD replay subset. In contrast, low or negative KER indicates that key experiences remain underrepresented in this subset, so a large total TD-error mass need not imply a large key-experience contribution. Consistent with this pattern, high test success rates are observed predominantly in configurations where both KER and KTC are high.

To quantify this dependence, we compute Spearman rank correlation coefficients between final success rate and KER at a converged training stage, denoted by $\rho$. In the Chemical environment, success rate exhibits a moderate positive correlation with KER $\rho = 0.37$, while correlations with learner participation ($K$) and replay coverage (UCR) are negligible ($|\rho| < 0.05$). In the Fire environment, this dependence is substantially stronger (SR--KER: $\rho = 0.70$), suggesting that performance is more sensitive to whether key experiences are represented in replay than to aggregation intensity or replay exposure alone.

High KTC alone, however, is not sufficient to ensure strong performance, particularly in the Fire environment. Because KTC is a relative measure within the high-TD replay subset, it does not capture the absolute availability or diversity of key experiences. Thus, even when key experiences account for a large share of TD-error mass within that subset, replay may still lack sufficient coverage of other learning-relevant transitions to support stable policy improvement.


Together, these results suggest that replay exposure influences performance by shaping both key experience admission and subsequent learning signal allocation.

\subsection{Performance and Energy Implications under Fixed Experience Budgets}
\begin{figure}[t]
  \centering
  \includegraphics[width=\columnwidth]{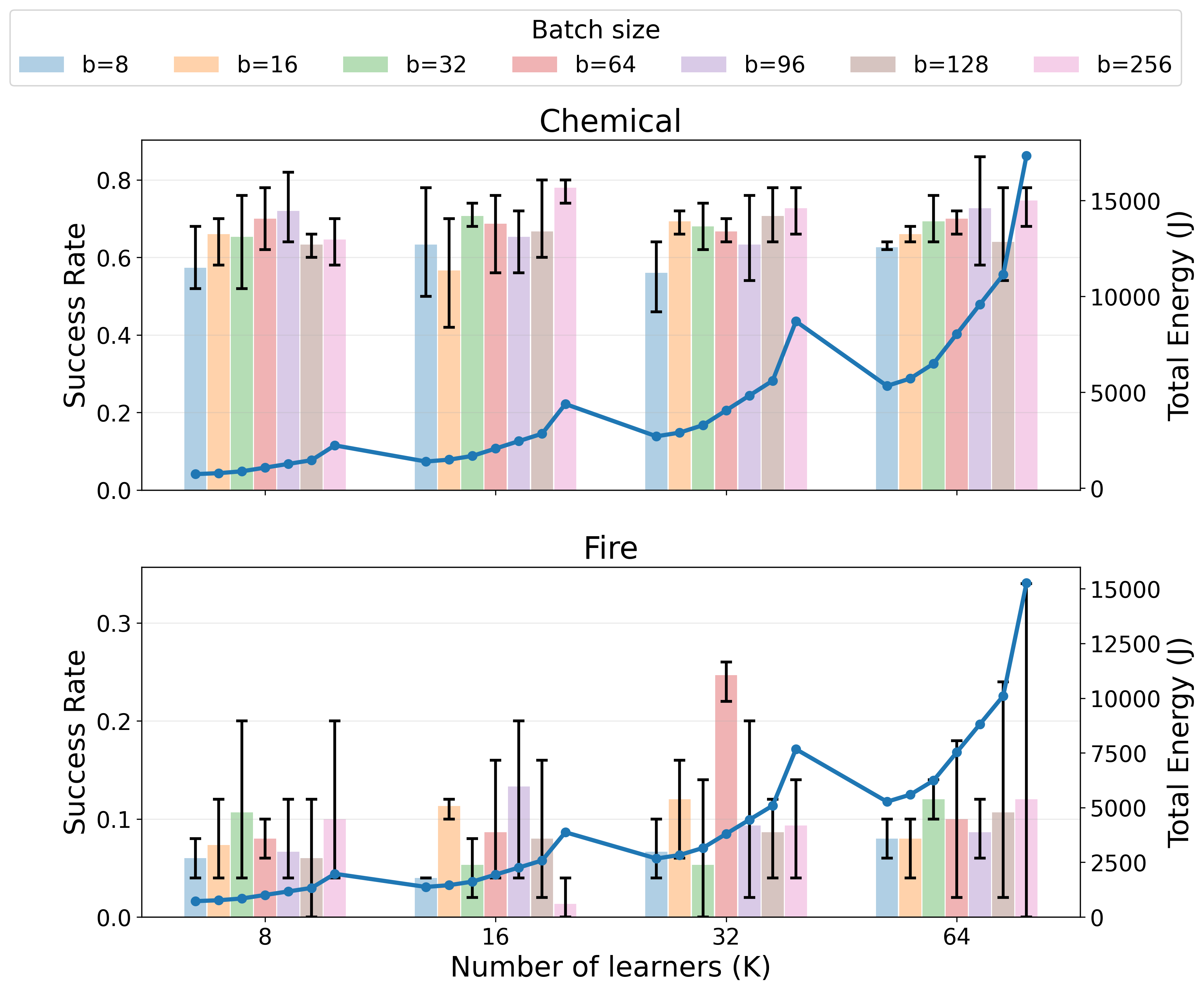}
  \caption{Performance--energy trade-off across different $(K,b)$ configurations. Bars represent success rate, while the blue curve indicates the corresponding system-level energy cost.}
  \label{fig:performance}
\end{figure}
We next evaluate task performance across learner participation levels and minibatch sizes to assess how the learning-signal patterns identified in Sec.~\ref{sec:sr} manifest at the outcome level. Performance is interpreted jointly with replay exposure and key experience admission under fixed per-round experience budgets.

Figure~\ref{fig:performance} reports test performance over the $(K,b)$ grid for both environments. Across tasks, minibatch structure is associated with clear and systematic performance trends, while learner participation primarily modulates performance magnitude without altering the underlying regime. 
When interpreted together with the diagnostics in Fig.~\ref{fig:ucr}, strong performance is predominantly observed in configurations with stronger key-experience representation in the high-TD replay subset, where key experiences dominate learning signal allocation. Conversely, configurations with limited key experience admission exhibit inferior performance, even when replay exposure or aggregate TD magnitude is comparatively high.

Under experience-constrained federated learning, changing minibatch size induces larger and more consistent variation in success rate than changing learner participation.
Specifically, across the $(K, b)$ configurations shown in Fig.~\ref{fig:ucr}, varying minibatch size induces success rate changes of approximately 10–18\%, whereas varying learner participation yields substantially smaller and less consistent differences within the decentralized aggregation regime.

We further examine the energy implications of learner participation to expose the performance--energy trade-off under fixed experience budgets.
As shown in Fig.~\ref{fig:performance}, configurations achieving comparable success rates can incur substantially different energy costs depending on learner participation.
In particular, increasing $K$ consistently raises communication and computation energy due to parallel optimization and aggregation, while providing limited or unstable performance gains once key experience admission saturates.
This reveals a trade-off: under constrained experience generation, higher participation primarily increases energy expenditure rather than improving learning effectiveness, whereas appropriately chosen $(K,b)$ configurations can achieve competitive and stable performance at significantly lower energy cost.

\vspace{-0.2em}
\section{Conclusion}

Under fixed experience budgets, increasing learner participation does not reliably improve performance and instead mainly increases experience reuse and energy cost. Our results show that performance in experience-constrained federated reinforcement learning is governed more by replay dynamics under limited data than by participation density or aggregation scope, clarifying when sparse participation is sufficient.

\section*{Acknowledgements}
This research is partially supported by the Air Force Office of Scientific Research (AFOSR), under contract FA9550-24-1-0078, and NSF award CNS-2148253. The paper was received and approved for public release by Air Force Research Laboratory (AFRL) on case number AFRL-2026-0974. Any opinions, findings, and conclusions or recommendations expressed in this material are those of the authors and do not necessarily reflect the views of AFRL or its contractors.

Portions of this manuscript were refined using an AI-assisted language model (ChatGPT, OpenAI) for grammar and clarity. The authors reviewed and edited the content to ensure correctness and originality.

\nocite{*} 
\bibliographystyle{IEEEtran}
\bibliography{ijcai26}

\end{document}